\documentclass[journal, twoside]{IEEEtran}
\usepackage[usenames]{color}
\usepackage{epsfig}
\usepackage{graphics}
\usepackage{amsmath,amssymb,amsfonts}
\usepackage{multirow}
\usepackage{cite}
\usepackage{array}
\usepackage{pslatex} 
\usepackage{url}
\usepackage{float}
\usepackage{caption, subcaption}
\usepackage{lineno}
\usepackage{graphicx}  
\usepackage{setspace}
\usepackage{tikz}
\usepackage{hhline}
\usepackage{mathtools}
\usepackage{algorithm}
\usepackage{algpseudocode}
\usepackage{comment}
\usepackage[letterpaper]{geometry}
 \newcommand{\flogo}{\includegraphics[height=0pt]{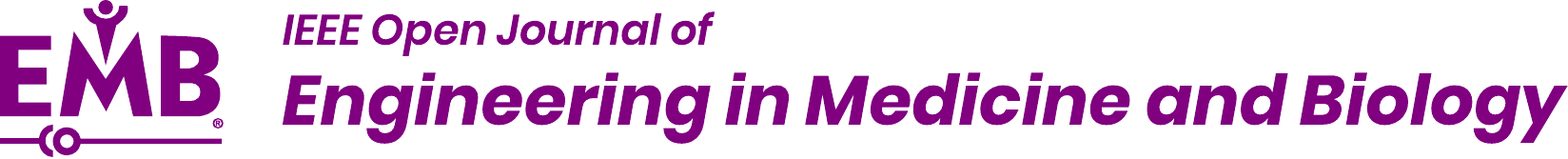}}
\ifCLASSINFOpdf
\else
\fi

\usepackage[english]{babel}
\usepackage[utf8]{inputenc}
\usepackage{fancyhdr}
\begin{document}
\title{\textcolor{violet}{Guided Conditional Diffusion Classifier (ConDiff) \\ for Enhanced Prediction of Infection in \\ Diabetic Foot Ulcers\vspace{0.25cm}}}
\author{Palawat~Busaranuvong, Emmanuel~Agu, Deepak~Kumar, Shefalika~Gautam, Reza~Saadati Fard, Bengisu~Tulu, Diane~Strong 
\thanks{This work is supported by the National Institutes of Health (NIH) through grant 1R01EB031910-01A1 Smartphone-based wound infection screener by combining thermal images and photographs using deep learning methods. (\textit{Corresponding author: Emmanuel Agu.})} 
\thanks{Palawat Busaranuvong and Shefalika Gautam are with the Data Science Department, Worcester Polytechnic Institute, Worcester, MA 01609 USA. Emmanuel Agu, Deepak Kumar, and Reza Saadati Fard are with the Computer Science Department, Worcester Polytechnic Institute, Worcester, MA 01609 USA. Bengisu Tulu and Diane Strong are with the Foisie Business School, Worcester Polytechnic Institute, Worcester, MA 01609 USA. 
(e-mail: pbusaranuvong@wpi.edu, emmanuel@wpi.edu, dkumar1@wpi.edu, sgautam@wpi.edu, rsaadatifard@wpi.edu, bengisu@wpi.edu, dstrong@wpi.edu).}}

\maketitle\thispagestyle{fancy}
\begin{abstract}
\textit{Goal:} To detect infected wounds in Diabetic Foot Ulcers (DFUs) from photographs, preventing severe complications and amputations.
\textit{Methods:} This paper proposes the Guided Conditional Diffusion Classifier (ConDiff), a novel deep-learning infection detection model that combines guided image synthesis with a denoising diffusion model and distance-based classification. The process involves (1) generating guided conditional synthetic images by injecting Gaussian noise to a guide image, followed by denoising the noise-perturbed image through a reverse diffusion process, conditioned on infection status and (2) classifying infections based on the minimum Euclidean distance between synthesized images and the original guide image in embedding space.
\textit{Results:} ConDiff demonstrated superior performance with an accuracy of 83\% and an F1-score of 0.858, outperforming state-of-the-art models by at least 3\%. The use of a triplet loss function reduces overfitting in the distance-based classifier.
\textit{Conclusions}: ConDiff not only enhances diagnostic accuracy for DFU infections but also pioneers the use of generative discriminative models for detailed medical image analysis, offering a promising approach for improving patient outcomes.
\end{abstract}

\begin{IEEEkeywords}
Diabetic Foot Ulcers, Diffusion Models, Distance-based Image Classification, Generative Models, Wound Infection.
\end{IEEEkeywords}

\IEEEpeerreviewmaketitle
\textbf{\textit{Impact Statement-} ConDiff enhances the accuracy of automatic diagnosis of infections in diabetic foot ulcers, offering a promising tool for early detection and improved patient care.}\\
\\

\section{Introduction}

\label{sec:introduction}

\IEEEPARstart{C}{hronic} wounds, affecting over 6.5 million people or approximately 2\% of the U.S. population, represent a significant health issue with healthcare expenses exceeding \$25 billion each year~\cite{jarbrinkhumanistic, sen2009human}. Among these, Diabetic Foot Ulcers (DFUs) are a prevalent subtype that poses a substantial risk for patients with diabetes. The commonality of DFUs on lower limbs, especially on the soles of the feet, makes them highly susceptible to infection~\cite{landis2008chronic}, with an alarming 40\% to 80\% of cases leading to infection~\cite{richard2011new}. These infections, often caused by bacteria, can result in severe complications, including cell death, limb amputation, and hospitalization~\cite{mills2014society}. Given these risks, effective monitoring and early detection of infections are crucial in the management of DFUs to prevent further complications.



\begin{table*}[h]
\centering
\caption{Summary of prior work on wound infection classification using deep learning}
\resizebox{\textwidth}{!}{
\begin{tabular}{cccccc}
\hline
\textbf{Specific ML problem} &
  \textbf{Related Work} &
  \textbf{\begin{tabular}[c]{@{}c@{}}Summary of \\ Approach\end{tabular}} &
  \textbf{\begin{tabular}[c]{@{}c@{}}No. of \\ Target Classes\end{tabular}} &
  \textbf{Dataset} &
  \textbf{Results} \\ \hline
\centering
\begin{tabular}[c]{@{}c@{}}Wound segmentation \\ and Infection \\ Classification\end{tabular} &
  \begin{tabular}[c]{@{}c@{}}Wang et al. \\ 2015 \cite{wang2015unified}\end{tabular} &
  \begin{tabular}[c]{@{}c@{}}CNN-based: \\ ConvNet + SVM\end{tabular} &
  \multirow{2}{*}{\begin{tabular}[c]{@{}c@{}}2 classes \\ (infection and\\  no infection)\end{tabular}} &
  \begin{tabular}[c]{@{}c@{}}NYU wound \\ Database\end{tabular} &
  \begin{tabular}[c]{@{}c@{}}Accuracy: 95.6\%\\ PPV: 40\%\\ Sensitivity: 31\%\end{tabular} \\ \cline{1-3} \cline{5-6} 
\centering
\begin{tabular}[c]{@{}c@{}}Classification of \\ 7 tissue types \\ including infection\end{tabular} &
  \begin{tabular}[c]{@{}c@{}}Nejati et al. \\ 2018 \cite{nejati2018fine} \end{tabular} &
  \begin{tabular}[c]{@{}c@{}}CNN-based: AlexNet\\ + PCA + SVM\end{tabular} &
   &
  \textbf{\begin{tabular}[c]{@{}c@{}}Private data\\ (data statistics \\ is unknown)\end{tabular}} &
  \textbf{\begin{tabular}[c]{@{}c@{}}Accuracy 95.6\%\\ (Only reported accuracy)\end{tabular}} \\ \hline
\centering
 &
  \begin{tabular}[c]{@{}c@{}}Goyal et al.\\ 2020 \cite{partb_DFU} \end{tabular} &
  \begin{tabular}[c]{@{}c@{}}CNN-based: \\ Ensemble CNN\end{tabular} &
  \multirow{3}{*}{\begin{tabular}[c]{@{}c@{}} \\ \\ 2 classes \\ (infection and \\ no infection)\end{tabular}} &
  \multirow{3}{*}{\textit{\begin{tabular}[c]{@{}c@{}} \\ Part B DFU \\  2020 dataset\\ \\ (We also used \\ this dataset)\end{tabular}}} &
  \begin{tabular}[c]{@{}c@{}}Accuracy: 72.7\%\\ PPV: 73.5\%\\ Sensitivity: 70.9\%\end{tabular} \\ \cline{2-3} \cline{6-6} 
\centering
\multirow{-2}{*}{\begin{tabular}[c]{@{}c@{}}DFU infection \\ classification\end{tabular}} &
  \begin{tabular}[c]{@{}c@{}}Al-Garaawi et al. \\ 2022 \cite{al2022diabetic} \end{tabular} &
  \begin{tabular}[c]{@{}c@{}}CNN-based:\\ DFU-RGB-TEX-Net\end{tabular} &
   &
   &
  \begin{tabular}[c]{@{}c@{}}Accuracy: 74.2\%\\ PPV: 74.1\%\\ Sensitivity: 75.1\%\end{tabular} \\ \cline{2-3} \cline{6-6} 
\centering
 &
   \begin{tabular}[c]{@{}c@{}}Liu et al. \\ 2022 \cite{liu2022diabetic} \end{tabular} &
  \begin{tabular}[c]{@{}c@{}}CNN-based:\\ augmentations\\ + EfficientNet\end{tabular} &
   &
   &
  \begin{tabular}[c]{@{}c@{}} \textbf{Data leakage} \\ when splitting \& \\  performing augmentations\end{tabular} \\ \hline
\centering
\multirow{2}{*}{\begin{tabular}[c]{@{}c@{}}DFU wound ischemia\\ and  infection \\ classification\end{tabular}} &
  \begin{tabular}[c]{@{}c@{}}Yap et al. \\ 2021 \cite{yap2021analysis} \end{tabular} &
  \begin{tabular}[c]{@{}c@{}}CNN-based:\\ VGG, ResNet,\\ InceptionV3, DenseNet, \\ EfficientNet\end{tabular} &
  \multirow{2}{*}{\begin{tabular}[c]{@{}c@{}}4 classes\\ (both infection\\ and ischemia,\\ infection, ischemia,\\ none)\end{tabular}} &
  \multirow{2}{*}{\begin{tabular}[c]{@{}c@{}} \\  DFUC2021\\ dataset\end{tabular}} &
   
  \begin{tabular}[c]{@{}c@{}}EfficientNet B0 \\ performance: \\ F1, PPV, SEN\\ =  55\% , 57\%, 62\%\end{tabular} \\ \cline{2-3} \cline{6-6} 
\centering
 &
  \begin{tabular}[c]{@{}c@{}}Galdran et al. \\ 2021 \cite{galdran2021convolutional} \end{tabular} &
  \begin{tabular}[c]{@{}c@{}}ViT-based: ViT,\\ DeiT, BiT\end{tabular} &
   &
   &
  \begin{tabular}[c]{@{}c@{}}BiT performance:\\ F1, PPV, SEN\\ = 61\%, 66\% , 61\%\end{tabular} \\ \hline
\end{tabular}
}
\label{tab:summary_of_approaches}
\end{table*}

\textbf{The problem}: Accurately diagnosing infections in DFUs involves analyzing the wound's bacteriology, and reviewing patient records including clinical history, physical health assessments, and blood tests. However, as clinicians do not always have access to this comprehensive wound information, they often rely on visual inspection to identify signs of infection in DFUs. Visual indicators of infection include increased redness around the ulcer and colored purulent discharge. Moreover, experienced wound experts are not always available, especially in low-resource settings and developing countries. This paper proposes an automated method that uses deep-learning to detect infected DFUs from images.  


\textbf{Prior work}: Recently, machine learning methods have achieved impressive performance in various medical image analysis and wound assessment tasks including works by Liu et al.~\cite{PWAT_cnn, PWAT_eff} for scoring the healing progress of chronic wounds from photographs based on evidence-based rubrics such as the Photographic Wound Assessment Tool (PWAT). Additionally, State-Of-The-Art (SOTA) machine learning image classification techniques have been proposed for detecting infections from the visual appearance of wounds in photographs~\cite{partb_DFU, galdran2021convolutional, yap2021analysis} without the need for direct wound tests, medical notes, or extensive clinical examinations. Goyal et al.~\cite{partb_DFU} introduced the CNN-Ensemble model, which extracts bottleneck features from Inception-V3, ResNet50, and InceptionResNetV2 that are then classified using an SVM classifier. Goyal et al.~\cite{partb_DFU} also described the DFU infection dataset, in which clinical experts at Lancashire Teaching Hospital in the United Kingdom, used visual inspection to label whether wound images are infected. CNN-Ensemble achieved 72.7\% accuracy for binary infection classification of wound images in the DFU infection dataset.

\begin{figure}[thb]
  \centering
  \includegraphics[width=0.8\linewidth]{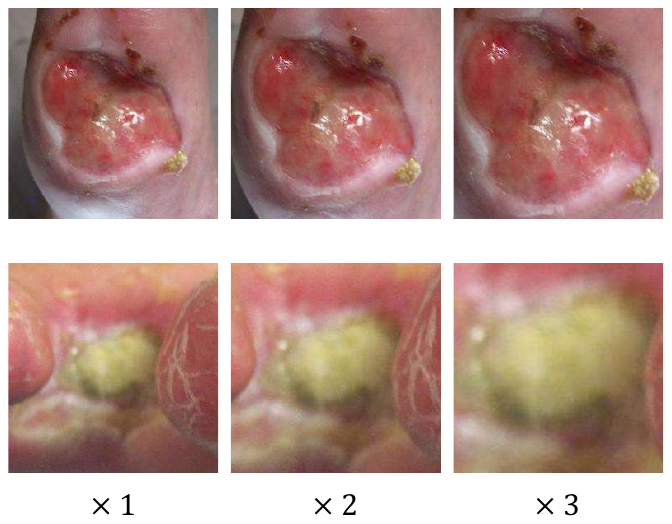}
  \caption{Natural data augmentation of an original image with three different magnifications.}
  \label{fig:magnificant}
\end{figure}

In a subsequent study, Liu et al.~\cite{liu2022diabetic} reported an impressive accuracy of 99\% for wound infection classification by adapting the EfficientNet model~\cite{efficientnet}, along with data augmentation techniques. However, their high accuracy was in part due to data leakage issues between the training and testing datasets. Specifically, the original DFU infection dataset from Goyal et al.~\cite{partb_DFU} included each wound image in three naturally augmented forms with varying magnifications (see Fig.~\ref{fig:magnificant}). Liu et al.~\cite{liu2022diabetic} randomly divided these augmented images between the training and testing sets on a sample-wise basis, which resulted in significant data leakage as the testing set included images that closely resembled augmented versions of the training images.

\textbf{Challenges}: Automated DFU image analyses to detect infection using deep learning methods face several challenges that hinder accurate diagnoses. Firstly, the distinction between images of infected and uninfected wounds is fine-grained~\cite{partb_DFU}, exemplified by high inter-class similarity and intra-class variation, which present a challenge to machine learning infection classifiers. Secondly, wound image datasets often have inconsistent imaging conditions, including variations in camera distance, orientation, and lighting~\cite{partb_DFU}. Lastly, as data collection in clinical environments is expensive and tedious, labeled DFU image datasets are often small, making it challenging to train deep-learning models.

\begin{figure}[t]
  \centering
    \includegraphics[width=1\linewidth]{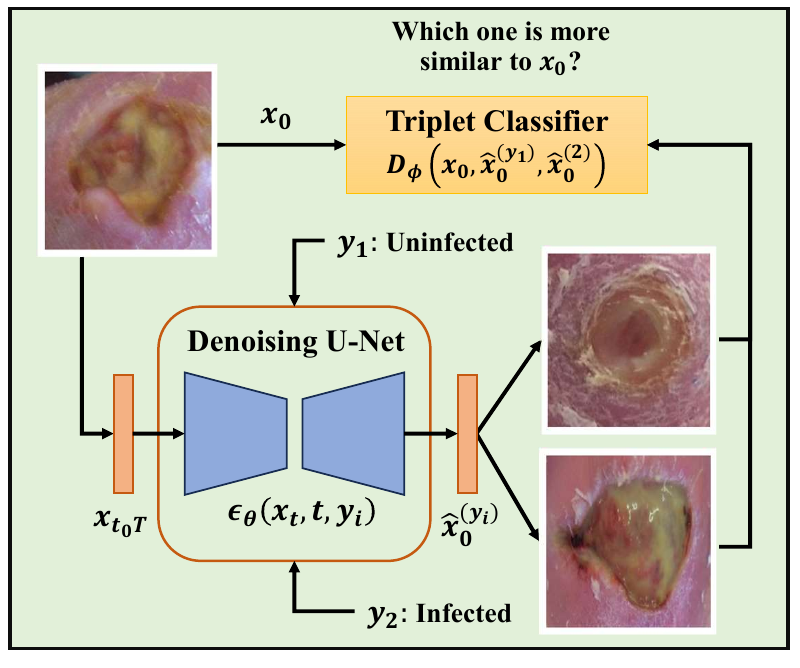}
  \caption{Inference in the ConDiff Classier Framework. Input $x_0$ is perturbed by noise of strength $t_0$. The perturbed input $x_{t_0 T}$ is denoised through a reverse diffusion process to synthesize image $\hat{x}_0^{(y_i)}$ conditioned on label $y_i$. Infection classification is based on the minimum $L_2$ distance between $x_0$ and $\hat{x}_0^{(y_i)}$ in embedding space.} 

  \label{fig:condiff_framework}
  \vspace{-4mm}
\end{figure}

\textbf{Our approach}: This paper presents the Guided Conditional Diffusion Classifier (ConDiff), a novel generative discriminative approach for wound infection classification (see  Fig.~\ref{fig:condiff_framework}). ConDiff leverages conditional guide image editing with a generative diffusion model~\cite{ stable_diffusion, sdedit} by perturbing an input image with a specific amount of Gaussian noise, and generating new images by using a reverse diffusion process to gradually remove noise from the noise-perturbed input image. ConDiff's diffusion process is conditioned on the state of the wound (no infection ($y_1$) or infection ($y_2$)), creating synthetic images reflective of these states. One key importance is the ability of ConDiff to discern and learn similarities between representations of the conditionally generated images $\hat{x}^{y}_{0}$ and the original wound image $x_0$ via an $L_2$ distance-based classifier in embedding space. The condition that yields the synthetic image that is most similar to the original, is selected as the predictive label. Unlike traditional supervised classification techniques that minimize a binary cross-entropy loss function, ConDiff mitigates overfitting by utilizing the triplet loss function~\cite{facenet}, to increase the distances between non-similar image pairs and reduce the distances between similar ones.
This work utilizes the DFU infection dataset provided by Goyal et al.~\cite{partb_DFU} (see Table.~\ref{tab:summary_of_approaches}). However, to eliminate data leakage between the training and test sets, we have refined our dataset creation and splitting strategy. Using subject-wise splitting, only the second magnified naturally augmented image (refer to Fig.~\ref{fig:magnificant}) is utilized for each subject.



\textbf{The main contributions} of this paper are: 

\begin{itemize}[]
    \item We propose the Guided Conditional Diffusion Classifier (ConDiff), an integrated end-to-end framework for classifying infected wound images. The ConDiff framework has 2 main components: (1) Guided diffusion, which injects Gaussian noise into a DFU image and then gradually removes noise from noise-perturbed images conditioned on infection status to synthesize conditional images (2) A distance-based classifier, which predicts an input image's label based on the minimum $L_2$ distance between original and synthesized images in embedding space. To the best of our knowledge, ConDiff is the first generative discriminative method to analyze fine-grained wound images, advancing the detection of Diabetic Foot Ulcer (DFU) infections.

  \item In rigorous evaluation on unseen test wound images (148 infected and 103 uninfected)  from the DFU infection dataset~\cite{partb_DFU}, our ConDiff framework significantly outperformed state-of-the-art baselines, improving on the accuracy and F1-score of wound infection detection by at least 3\%. 
  
  \item We demonstrate that by minimizing the Triplet loss function during training, ConDiff reduces overfitting on a small DFU  dataset of 1416 training images.  
    
  \item Heatmaps generated by Score-CAM \cite{score_cam} are applied to visually illustrate that ConDiff focuses on the correct wound image regions when classifying wound infection status.

\end{itemize}


\section{Related Work}
\label{sec:RelatedWork}

\subsection{Wound Infection Classification with Deep Learning}
Two neural network architectures have predominantly been popular for image analyses: Convolutional Neural Network (CNN) and Vision Transformer (ViT) based models. Table.~\ref{tab:summary_of_approaches} summarizes prior work on DFU infection classification from wound images. Unlike other prior discriminative approaches that directly predict the probability of each label from wound images, ConDiff utilizes a generative discrimination framework to predict a wound image's infection label based on the guided synthetic image that has the highest similarity to the input image in embedding space. 
\subsection{Generative Diffusion Models}

\textit{Diffusion Models}~\cite{DDPM, DDIM} have recently emerged as a new class of generative models, drawing attention for their robustness and high-quality outputs, and presenting an alternative to Generative Adversarial Networks (GANs). In contrast to the adversarial approach used to train GANs, diffusion models incrementally learn to reverse a process that introduces noise into data. This method avoids issues with GANs such as mode collapse, yielding a significantly more stable training process. Diffusion models have shown great potential in various applications including Text-to-Image generation and Image-to-Image translation~\cite{stable_diffusion, controlnet, nichol2021glide}. 
A standout quality of diffusion models is their ability to generate highly detailed and varied outputs, which is particularly advantageous in medical imaging. Recent studies have demonstrated the potential of diffusion models to effectively synthesize medical images including detecting and segmenting brain anomalies~\cite{wyatt2022anoddpm, wu2022medsegdiff}
, performing multi-contrast MRI and MRI-CT translations~\cite{ozbey2023unsupervised}, and enhancing the resolution of 3D Brain MRIs~\cite{wang2023inversesr}. To the best of our knowledge, diffusion models have not previously been applied to wound image analyses.

\section{Methodology}

\label{sec:methodology}





We now expound on the ConDiff framework, its components, theoretical bases, and practical application. 

\subsection{Denoising Diffusion Probabilistic  Model (DDPM)}
\label{sec:ddm}
DDPM~\cite{DDPM} is a generative model that leverages diffusion processes to generate synthetic data. DDPM has two main stages: 1) a forward process and 2) a reverse process. The forward process $q(x_{1:T}|x_0)$ systematically adds Gaussian noise to the data in $T$ steps, ultimately transforming the original data into a Gaussian distribution $p(z)$. In the reverse process $p_\theta(x_{0:T})$, the model learns to remove this noise iteratively to reconstruct or generate data samples, accomplished using a neural network trained to predict and reverse the noise added at each step. 


\subsubsection{Forward Diffusion Process}
\label{subsec:forward_ddpm}

Starting with an initial data point $x_0 \sim q(x)$, to which noise is incrementally added over $T$ steps. At each step $t$, this process is represented as a Gaussian transition in a Markov chain:
\begin{align}
x_t &= \sqrt{1-\beta_t}x_{t-1}+\sqrt{\beta_t}\epsilon_t
\nonumber \\
q(x_t|x_{t-1}) &= \mathcal{N}(x_t; \mu_t=\sqrt{1-\beta_t}x_{t-1}, \Sigma_t=\beta_t I)
\label{eq:markov1}
\end{align}

where $\epsilon_t \sim \mathcal{N}(0,I)$ represents the Gaussian noise, and $\beta_t < 1$ denotes the variance schedule. Considering $\alpha_t=1-\beta_t$ and $\Bar{\alpha}_t=\prod_{\tau=1}^{T}{\alpha_\tau}$, a tractable closed-form for sampling at any step $t$, given $x_0$ can be defined as (Eq.~\ref{eq:xt_eq}).
\begin{align}
q(x_t|x_0) &= \mathcal{N}(x_t; \sqrt{\Bar{\alpha}_t}x_0, (1-\Bar{\alpha}_t)I)
\label{eq:xt_eq}
\end{align}

The entire forward diffusion process can be viewed as a joint posterior distribution: \textbf{$q(x_{1:T}|x_0) = \prod_{t=1}^{T}{q(x_t|x_{t-1})}$}.

\subsubsection{Reverse Diffusion Process}
Since the distribution $q(x_0)$ of the underlying data is unknown, analytically reversing the forward diffusion process to get $q(x_{t-1}|x_t)$, is intractable. To manage this, the reverse process is approximated using a neural network-based distribution $p_\theta(x_{t-1}|x_t)$, characterized by learnable parameters $\theta$. The reverse process' joint distribution is given by $p_\theta(x_{0:T}) = p_\theta(x_{T})\prod_{t=1}^{T}{p_\theta(x_{t-1}|x_t)}$.
Here, the reverse process starts from a Gaussian distribution $p(x_T)$ and follows learned Gaussian transitions defined as:
\begin{equation}
p_\theta(x_{t-1}|x_t)=\mathcal{N}(x_{t-1}; \mu_\theta(x_t,t), \Sigma_\theta(x_t,t))
\label{eq:sampling_dist}
\end{equation}

where $\mu_\theta$ and $\Sigma_\theta$ are the neural network's predictions for the mean and variance at each step $t$.

\subsubsection{Learning Objective}

Due to the intractability of exact likelihood computation in this context, a technique adopted from Variational Autoencoders (VAEs) \cite{vae} is utilized, focusing on maximizing the Evidence Lower Bound (ELBO) \cite{bishop2023deep}.
\begin{align}
L(\theta) &= \mathbb{E}_{q}\left[\log\frac{p_\theta(x_{0:T})}{q(x_{1:T}|x_0)}\right] \nonumber \\
&= \mathbb{E}_{q}\Bigg[ - \sum_{t>1}D_{KL}\left(q(x_{t-1}|x_t) \| p_\theta(x_{t-1}|x_t)\right) \nonumber  \\
&\qquad\qquad + \log p_\theta(x_0|x_1) \Bigg] \label{eq:elbo}
\end{align}

Ho et al.~\cite{DDPM} further refined the objective function in Eq.~\ref{eq:elbo} to yield a simplified objective function for minimizing the Mean Squared Error (MSE) between the actual and predicted noise (Eq.~\ref{eq:mse_object}).
\begin{align}
L_{DM}(\theta) &= \mathbb{E}_{x_0,t,\epsilon\sim \mathcal{N}(0,I)}\left[ {\|\epsilon-\epsilon_\theta(x_t,t)\|}^2_2 \right]
\label{eq:mse_object}
\end{align}

 Here, $\epsilon_\theta(x_t,t)$ is a denoising U-Net model used for approximating $\epsilon$ from $x_t$, which aids in determining the mean $\mu_\theta(x_t, t)$:
\begin{equation}
\mu_\theta(x_t, t) = \sqrt{\Bar{\alpha}_{t-1}} F_\theta(x_t,t) +
\sqrt{1-\Bar{\alpha}_{t-1}-\sigma^{2}_t} \epsilon_\theta(x_t,t)
\label{eq:mu_theta}
\end{equation}
\begin{align}
\text{where } F_\theta(x_t,t) &=\frac{x_t-\sqrt{1-\Bar{\alpha}_t}\epsilon_\theta(x_t,t)}{\sqrt{\Bar{\alpha}_{t}}} \nonumber \\
\text{and } \sigma^2_t &= \frac{1 - \bar{\alpha}_{t-1}}{1 - \bar{\alpha}_t} \beta_t
\label{eq:mean_eq}
\end{align}

\subsubsection{Sampling Process with DDIM}
\label{subsec:ddim}

Generating new samples with DDPM is computationally intensive, often requiring about a thousand sampling steps. To address this, Song et al.~\cite{DDIM} introduced the Denoising Diffusion Implicit Model (DDIM), which accelerates the sampling process by significantly reducing the number of iterative steps. This is achieved by redefining the diffusion process as a non-Markovian process. 
Considering the DDPM sampling process in Eq.~\ref{eq:sampling_dist}, the latent $x_{t-1}$ can be derived from $x_t$ using the parameterized mean $\mu_\theta(x_t,t)$ in Eq.~\ref{eq:mu_theta} and 
variances $\Sigma_\theta(x_t,t)=\sigma^2_t I$.
\begin{equation}
x_{t-1} = \sqrt{\Bar{\alpha}_{t-1}} F_\theta(x_t,t) +
\sqrt{1-\Bar{\alpha}_{t-1}-\sigma^{2}_t} \epsilon_\theta(x_t,t) + \sigma_t z
\label{eq:ddpm_sampling}
\end{equation}

Here, $z \sim \mathcal{N}(0,I)$ represents standard Gaussian noise, independent of $x_t$. In DDIM, setting $\sigma_t=0$ for all $t$ makes the process deterministic, avoiding the need for a Markov chain and allowing for step-skipping. Eq.~\ref{eq:ddm_sampling} expresses the DDIM sampling step. This approach significantly enhances efficiency, generating new samples $10 \times$ to $50 \times$ faster than DDPM.
\begin{equation}
x_{t-1} = \sqrt{\Bar{\alpha}_{t-1}} F_\theta(x_t,t) +
\sqrt{1-\Bar{\alpha}_{t-1}} \epsilon_\theta(x_t,t)
\label{eq:ddm_sampling}
\end{equation}

\subsection{Conditional Image Generation with Diffusion Models}
\label{sec:Conditional Image Generation}

Prior works have demonstrated that these models can be adapted to conditional generation, where the generated data depends on a given condition or context, such as a class label or a textual description~\cite{Classifier_guidance, CFG}.
The simplest implementation method involves introducing the conditioning variable $y$ as an additional input to the denoising network, represented as $\epsilon_\theta(x_t,t,y)$. However, one limitation is that the network may not adequately consider the conditioning variable, sometimes completely overlooking it~\cite{bishop2023deep}. To address this, a \textit{guidance scale} is introduced, enhancing the influence of the conditioning variable during sample generation.

\subsubsection{Classifier-Free Diffusion Guidance}

Dhariwal and Nichol \cite{Classifier_guidance} introduced \textit{Classifier Guidance} where the diffusion score 
$\tilde{\epsilon}_\theta(x_t,t,y) = \nabla_{x_t} \log{p(x_t)} + \omega \nabla_{x_t} \log{p(y|x_t)}$ 
includes a classifier $p_\phi(y|x_t)$ that must be able to predict the samples with varying degrees of noise $x_t$. A guidance scale $\omega$ controls the weight given to the classifier gradient. To prevent the inclusion of the auxiliary classifier $p_\phi(y|x_t)$, Ho et al. \cite{CFG} later applied  Bayes’ theorem to reformulate $\nabla_{x_t} \log{p(y|x_t)}$. Consequently, the \textit{Classifier-Free Guidance (CFG)} formula is expressed as:
\begin{equation}
\tilde{\epsilon}_\theta(x_t,t,y) = (1-\omega)\epsilon_\theta(x_t,t) + \omega \epsilon_\theta(x_t,t,y) 
\label{eq:CFG_eq}
\end{equation}

To train this guided diffusion model, the label $y$ is incorporated as an additional input in the learning objective (Eq.~\ref{eq:mse_object}), resulting in Eq.~\ref{eq:mse_object2}.
\begin{equation}
L_{DM}(\theta) = \mathbb{E}_{x_0,t,y,\epsilon \sim \mathcal{N}(0,I)}\left[ {\|\epsilon-\epsilon_\theta(x_t,t,y)\|}^2_2 \right]
\label{eq:mse_object2}
\end{equation}

\subsection{Guided Conditional Diffusion Classifier (ConDiff) }
\label{sec:ConDiff}

To complete our ConDiff model, two additional components are now introduced: 1) guided image synthesis and 
2) triplet loss for learning similarity.

\subsubsection{Guided Image Synthesis}
\label{sec:SDEdit}
\begin{figure}[htbp]
\centering
  \includegraphics[width=\linewidth]{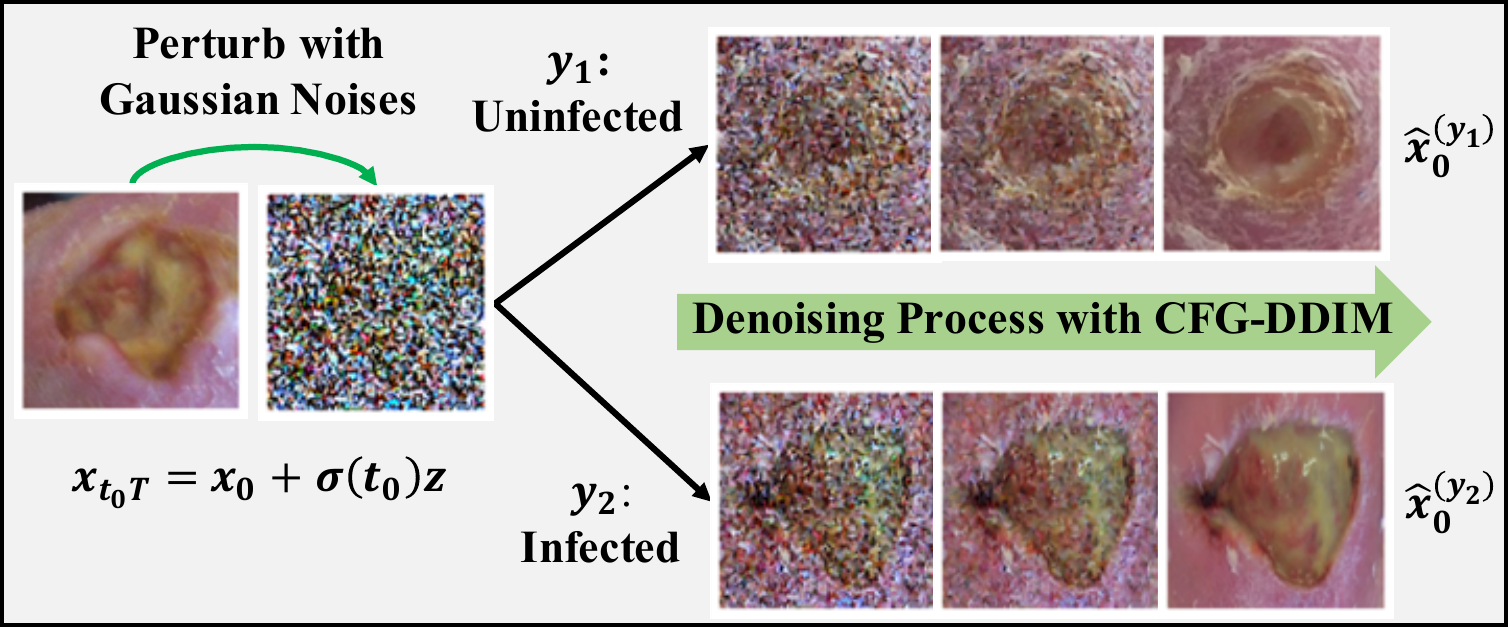}
  \caption{Synthesizing conditional DFU images using ConDiff. A guide image $x_0$ is perturbed with Gaussian noises that are then removed progressively using a CFG-DDIM sampling technique (see Algorithm.~\ref{alg_cfg_sampling}), conditioned on the infection status. This process gradually projects $x_0$ to guided synthetic images of conditions: $\hat{x}_0^{(y_1)}$ and $\hat{x}_0^{(y_2)}$.}
  
  \label{fig:ConDiff_gen}
  \vspace{-3mm}
\end{figure}


We aim to generate conditional images that are guided by original images. Specifically, we seek to synthesize wound images that closely resemble the input image, while also being distinct enough to differentiate between infection and non-infection conditions. We employ a strategy from image synthesis with Stochastic Differential Equations (SDEs) \cite{sdedit}, introducing a certain amount of Gaussian noise to the guide image through a forward diffusion process (see Eq.~\ref{eq:xt_eq}). 
Next, we synthesize conditional images using classifier-free guidance (Sec.~\ref{sec:Conditional Image Generation}). The noise strength $t_0 \in (0,1]$ indicates the level of noise added to the original image $x_0 \sim q(x)$:
\begin{equation}
x_{t_0T} = x_0 + \sigma(t_0)z, \text{ where } z \sim \mathcal{N}(0,I)
\label{eq:guide image}
\end{equation}
Here, $\sigma(t_0)=\sqrt{\frac{1-\Bar{\alpha}_{t_0 T}}{\Bar{\alpha}_{t_0 T}}}$ is a scalar function determining the noise magnitude, and $T$ is the total number of forward diffusion steps. Subsequently, the classifier-free guided DDIM sampling is used to synthesize images, as illustrated in Algorithm~\ref{alg_cfg_sampling} and visualized in Fig.~\ref{fig:ConDiff_gen}.  

Algorithm~\ref{alg_cfg_sampling} outlines ConDiff Sampling with CFG-DDIM. In line (1), Gaussian noise of strength $t_0$ is added to the input image. The reverse process begins from step $t_0T$ to step $1$ (line 2). Line (3) applies the modified diffusion model $\tilde{\epsilon}_\theta(x_t,t,y)$ following CFG (Eq.~\ref{eq:CFG_eq}), and the DDIM sampling process is conducted in line (4). 


%
\begin{algorithm}[htb]
\caption{ConDiff Sampling with CFG-DDIM}
\label{alg_cfg_sampling}
\begin{algorithmic}[1]
\Require Guide image: \( x_0 \), class label: \( y \), guidance scale: \( \omega \), noise strength: \( t_0 \), and number of diffusion steps: \( T \).
\State \( x_{t_0 T} = x_0 + \sigma(t_0)z, \) where \( z\sim\mathcal{N}(0, I) \)
\For{\( t = t_0T, \ldots, 1 \)}
    \State \( \tilde{\epsilon}_\theta(x_t,t,y) = \omega\epsilon_{\theta}(x_t, t, y) + (1 - \omega)\epsilon_{\theta}(x_t, t) \)
    \State \( x_{t-1} = \sqrt{\Bar{\alpha}_{t-1}} \left( \frac{x_t-\sqrt{1-\Bar{\alpha}_t}\tilde{\epsilon}_\theta}{\sqrt{\Bar{\alpha}_t}} \right) + \sqrt{1-\Bar{\alpha}_{t-1}} \tilde{\epsilon}_\theta
    \)
\EndFor
\State \textbf{return} \( \hat{x}_0^{(y)} \)
\end{algorithmic}
\end{algorithm}



\subsubsection{Learning Similarity with Triplet Loss}
Having generated images with different labels, our next goal is to determine which synthesized image is most similar to the guide image $x_0$, a sub-problem addressed using the triplet loss function (Eq.~\ref{eq:triplet_loss}) introduced in FaceNet~\cite{facenet}. The triplet loss function minimizes the distance between an anchor image $x^{(a)}$ and a positive image $x^{(p)}$ with the same identity while maximizing the distance between the anchor image and a negative image $x^{(n)}$ with a different identity. A neural network $f_\phi(x)$ transforms the images into a $d$-dimensional Euclidean space, facilitating similarity measurement.
\begin{equation}
\begin{aligned}
L_{triplet} = \mathbb{E}\bigg[ & \left(\|f_\phi(x^{(a)}) - f_\phi(x^{(p)})\|^2_2 \right. \\
& \left. - \|f_\phi(x^{(a)}) - f_\phi(x^{(n)})\|^2_2 + \alpha \right)_+ \bigg]
\end{aligned}
\label{eq:triplet_loss}
\end{equation}

The margin $\alpha$ is set to 1, indicating the desired separation between similar and dissimilar pairs. Consequently, our classifier $D_\phi$ determines the closest conditional synthesized image to the guide image $x_0$ by comparing their $L_2$ distance in embedding space (Eq.~\ref{eq:d_phi}).
\begin{equation}
D_\phi(x_0,\hat{x}_0^{(y_1)},\hat{x}_0^{(y_2)}) = \arg \min_{y_i} \left \{L_2 (f_\phi(x_0),f_\phi(\hat{x}_0^{(y_i)}) ) \right \}
\label{eq:d_phi}
\end{equation}

\subsubsection{ConDiff Training Process}
\label{sec: train condiff}
The training process is divided into two stages. Stage 1 involves training $\epsilon_\theta(x_t,t,y)$ with the objective expressed in Eq.~\ref{eq:mse_object2}, focusing on noise removal (Algorithm~\ref{alg_train_stage1}). After optimizing $\epsilon_\theta(x_t,t,y)$, conditional images $\hat{x}_0^{(y_1)}$ and $\hat{x}_0^{(y_2)}$ are generatd using ConDiff, following the process in Fig.~\ref{fig:ConDiff_gen} for all DFU images in our dataset. Stage 2 optimizes the embedding network $f_\phi$ to learn the Triplet loss (Eq.~\ref{eq:triplet_loss}, Algorithm~\ref{alg_train_stage2}).

\begin{algorithm}[h]
\caption{Jointly training a diffusion model with CFG}
\label{alg_train_stage1}
\begin{algorithmic}[1]
\Require diffusion model: \(\epsilon_\theta \), pairs of guide image and class label: \( q(x_0, y) \), probability of unconditional training: \( p_{\text{uncond}} \), and number of training steps: \(N \).
\Ensure learned \( \epsilon_\theta \)
\For{\( n = 1, \ldots, N \)}
    \State Sample a pair of image-label \( (x_0, y) \sim q(x_0, y) \)
    \State \( y \gets \emptyset \) with probability \( p_{\text{uncond}} \)
    \State \( t \sim \text{Uniform}\{1, \ldots, T\} \), \( \epsilon \sim \mathcal{N}(0, I) \)
    \State \( x_t = \sqrt{\Bar{\alpha}_t}x_0 + \sqrt{1 - \Bar{\alpha}_t}\epsilon \) 
    \State Take gradient step on \( \nabla_{\theta} \left\| \epsilon - \epsilon_\theta(x_t, t, y) \right\|^2 \)
\EndFor
\end{algorithmic}
\end{algorithm}

\begin{algorithm}[h]
\caption{Training a distance-based classifier model with Triplet loss}
\label{alg_train_stage2}
\begin{algorithmic}[1]
\Require embedding model: \( f_\phi \), real dataset: \( D_r \), synthetic dataset: \( D_s \), probability of sampling from \( D_s \): \( p_{gen} \), number of epochs: \( E \), and batch size: \( B \).
\Ensure learned \( f_\phi \)
\For{\( \text{epoch} = 1, \ldots, E \)}
    \For{\( \text{batch} = 1, \ldots, \lceil \text{size}(D_r)/B \rceil \)}
        \State \( (x^{(a)}, x^{(p)}, x^{(n)}) \sim D_r \) (sample \( B \) triplets)
        \State \( (x^{(p)}, x^{(n)}) \sim D_s \) with probability \( p_{gen} \)
        \State Compute \(L_{triplet}\) in Eq.~\ref{eq:triplet_loss}.
        \State Take gradient step on \( \nabla_\phi L_{triplet} \)
    \EndFor
\EndFor
\end{algorithmic}
\end{algorithm}



\underline{Training Stage 1}: Algorithm~\ref{alg_train_stage1} details joint training of the unconditional and conditional models. Occasionally, $y$ is set to the unconditional class identifier $\emptyset$ with a probability $p_{uncond}$, a hyperparameter typically around 10–20\% \cite{CFG}.  
For each training step, an image $x_0$ and condition $y$ are sampled from the dataset. The diffusion step $t$ is also sampled from the uniform distribution of $[1, \ldots ,T]$. Subsequently, in line (5), a forward diffusion process is performed to add noise to the image $x_0$ following Eq.~\ref{eq:xt_eq}. The diffusion model parameters $\theta$ are then optimized to minimize the MSE between the actual and predicted noise.

\underline{Training Stage 2}: The embedding model $f_\phi$ is trained in Algorithm~\ref{alg_train_stage2}. In addition to the real DFU dataset $D_r$, a synthetic dataset $D_s$ is also created from our trained ConDiff model following the sampling process in Algorithm~\ref{alg_cfg_sampling}. For each iteration, a batch of triplets $(x^{(a)}, x^{(p)}, x^{(n)})$ is sampled from $D_r$. To increase recognition ability and improve training robustness, $(x^{(p)}, x^{(n)})$ is sampled from $D_s$  with the probability of $p_{gen}=0.2$. Embedding model parameters $\phi$ are optimized to minimize the Triplet loss, thus learning to distinguish between similar and dissimilar images effectively.

\section{Experiments and Results}
\label{sec:experiments}
\subsection{Diabetic Foot Ulcer (DFU) Dataset}

The DFU Infection Dataset curated by Goyal et al.~\cite{partb_DFU}, comprises DFU images collected at the Lancashire Teaching Hospital with the permission of the  U.K. National Health Service (NHS). These DFU images were labeled by two DFU specialists (consultant physicians) based on visual inspection, independent of medical notes or clinical tests. The dataset consists of 2,946 augmented patches with infection and 2,946 augmented patches without infection. Natural augmentation was performed with varying magnification as illustrated in Fig.~\ref{fig:magnificant}. All patches have dimensions of $224 \times 224 \times 3$ pixels. 




\textbf{Data Preprocessing}: The training, validation and test sets were created using 65\%, 15\% and 20\% of the dataset respectively. To prevent data leakage, subject-wise splitting was utilized, where all images for each case could only belong to one class. As a further measure to prevent data leakage, as shown in Fig.~\ref{fig:magnificant}, only augmented patches with a $\times 2$ magnification level were utilized. The model achieving the highest accuracy on the validation set (optimal parameter values), was selected for final evaluation on the test (unseen) dataset. Table.~\ref{tab:dataset_statistic} shows the dataset statistics after pre-processing. 

\begin{table}[thbp]
    \renewcommand{\arraystretch}{1.3}
    \caption{Refined DFU Dataset Statistics}
    \label{tab:dataset_statistic}
    \centering
    \begin{tabular}{ccc}
        \hline
        Processed Data & Category & \# of Patches \\ \hline
        \multirow{2}{*}{Train Data (65\%)} & Infection & 709 \\
        \cline{2-3}
        & No Infection & 495 \\
        \hline
        \multirow{2}{*}{Validation Data (15\%)} & Infection & 125 \\
        \cline{2-3}
        & No Infection & 87 \\
        \hline
        \multirow{2}{*}{Test Data (20\%)} & Infection & 148 \\
        \cline{2-3}
        & No Infection & 103 \\
        \hline
    \end{tabular}
\end{table}

\vspace{-5mm}
\subsection{Experimental Setup}
\subsubsection{Implementation Details}

As outlined in Sec.~\ref{sec: train condiff}, the ConDiff classifier was trained in two distinct stages on an NVIDIA A100 GPU. 

\underline{Stage 1 - Fine-Tuning the Diffusion Model}:
First, a Stable Diffusion model pre-trained on the LAION-400M dataset ~\cite{schuhmann2021laion}, a large collection of 400 million image-text pairs, was retrieved~\cite{stable_diffusion}. Its denoising U-Net $\epsilon_\theta$ was fine-tuned on our wound training set $D_r$, as described in Algorithm~\ref{alg_train_stage1}. The training involved 10,000 iterations, employing the AdamW optimizer with a learning rate of $1 \times 10^{-5}$ for fine-tuning. After training, our ConDiff generator was utilized to synthesize conditional DFU images with the following hyperparameter setting: guidance scale $\omega=0.75$, noise strength $t_0=0.8$, and the number of sampling steps $T=30$. These images were synthesized using training images as guide inputs, resulting in a collection of synthetic data $D_s$, for use in the second stage of training. 


\underline{Stage 2 - Training the Embedding Network $f_\phi$}: utilized the EfficientNet-B0 neural networks architecture  as the backbone. The Euclidean embedding dimension $d$ was set to 256. 
The optimal set of parameters $\phi$ was obtained from the best model performance on the validation partition. The training was for 50 epochs as described in Algorithm~\ref{alg_train_stage2}, using the AdamW optimizer with a learning rate of $1 \times 10^{-3}$.   

\subsubsection{Evaluation Metrics} of all models on the DFU dataset were as follows. 

\underline{Classification metrics}: Accuracy, sensitivity, specificity, Positive Predictive Value (PPV), and F1-score were used to evaluate DFU infection classification performance.  

\underline{Image generation metrics}: To evaluate the quality of synthetic images, the Fréchet Inception Distance (FID) score and the Inception Score (IS) were employed. A lower FID indicates more realism and higher quality while a higher IS indicates that the generated images are distinct and diverse.

\underline{Clustering metric}: DFU images' embedding vectors of uninfected and infected wounds were analyzed using the Silhouette score. It measures how similar data points within the cluster (cohesion) are compared to data points in other clusters (separation). Its value ranges from $-1$ to $1$, where a high value indicates that the object is well-matched to its cluster and poorly matched to neighboring clusters.

\subsubsection{SOTA Baseline Models}
\label{sec: baseline}
Recent deep learning image classification architectures including CNN and ViT-based models were considered as baselines. Due to the small size of our dataset, the base or tiny version of each model was selected for evaluation.

\begin{table*}[!ht]
\centering 
\caption{Quantitative comparison of ConDiff with SOTA baseline models on test images in the DFU infection dataset}
\label{tab:SOTA_comparison} 
\begin{tabular}{clcccccc}
\hline
\multicolumn{2}{c}{Model}  & Accuracy & F1-score & Sensitivity & Specificity & PPV & Test time  \\ \hline
\multirow{4}{*}{\begin{tabular}[c]{@{}c@{}}Convolutional\\ Neural\\  Networks\end{tabular}} 
& ResNet-18          & 0.785    & 0.817    & 0.817       & 0.738       & 0.818 & 1.01 sec.\\
& DenseNet-121       & 0.733    & 0.763    & 0.730       & 0.738       & 0.800 & 4.02 sec.\\
& Inception-V3       & 0.733    & 0.765    & 0.736       & 0.728       & 0.796 & 2.76 sec.\\
& EfficientNet-B0    & 0.793    & 0.810    & 0.750       & \textbf{0.854}       & \textbf{0.881} & 2.01 sec.\\ \hline
\multirow{4}{*}{\begin{tabular}[c]{@{}c@{}}Vision\\ Transformers\end{tabular}}
& ViT-Small          & 0.773    & 0.801    & 0.777       & 0.767       & 0.827 & 1.26 sec.\\
& DeiT-Tiny          & 0.789    & 0.828    & 0.824       & 0.738       & 0.819 & 1.25 sec.\\
& SwinV2-Tiny        & 0.785    & 0.822    & \textbf{0.878}     & 0.650      & 0.783 & 2.51 sec.\\
& EfficientFormer-L1 & \underline{0.801}    & \underline{0.833}    & 0.845   & 0.738       & 0.822  & 1.51 sec.\\ \hline
Diffusion
& \textbf{ConDiff (ours)}     & \textbf{0.833}    & \textbf{0.858}    & \underline{0.858}       & \underline{0.796}       & \underline{0.858}  & 12.55 mins \\ \hline
\end{tabular}

\end{table*}

\underline{CNN-based models}: ResNet~\cite{resnet} and Inception-V3~\cite{inception} were selected as baseline models because Goyal et al. \cite{partb_DFU} employed them as backbones in the ensemble CNN model for DFU infection classification. DenseNet~\cite{densenet} was selected as Yap et al. \cite{yap2021analysis} found that it achieved the best macro-F1 score in 4-class DFU image classification. 
EfficientNet~\cite{efficientnet} was selected as it was the most effective CNN-based model in analyzing wound infections \cite{galdran2021convolutional, yap2021analysis}.

\underline{ViT-based models}: ViT~\cite{vit} and DeiT~\cite{deit} were explored for for DFU ischemia \& infection classification by Galdran et al. \cite{galdran2021convolutional}, achieving a macro-F1 score comparable to the best CNN-based model (EfficientNet). SwinV2~\cite{swinv2} and EfficientFormer~\cite{efficientformer} were selected as baselines for DFU infection classification because
 even have not previously been explored for infection classification, they are recent architectures that outperformed previous ViT-based \& CNN-based models on ImageNet classification.


\subsubsection{Score-Weighted Class Activation Mapping (Score-CAM)}~\cite{score_cam} is a method for interpreting the decision-making process of CNN models in visual tasks. It creates a visual heatmap that reveals which regions of a conditional synthesized image $\hat{x}_0^{(y)}$ the embedding model $f_\phi$ found similar to a guide (input) image $x_0$. Score-CAM has the following steps: (1) extracting feature maps $A^{k}$ from the last convolutional layer $L$; $A^{k}=f_{\phi,L}(\hat{x}_0^{(y)})$, (2) generating activation maps that emphasize predictive regions of the image; $M^{k}=\text{Upsample} (A^{k})$, (3) calculating cosine similarity scores between the guide image embedding and the corresponding synthesized images based on the activation maps; $S^{k}=\text{Similarity}(f_\phi(x_0), f_\phi(M^{k} \circ \hat{x}_0^{(y)}))$, and (4) aggregating these activation maps into a comprehensive heatmap, with the weighting determined by the similarity scores $\alpha^k=\frac{\exp(S^k)}{\Sigma_{j}{\exp(S_j^k)}}$; $H_{\text{Score-CAM}}=\text{ReLU}(\Sigma_{k}{\alpha^{k}M^{k}})$.

\subsection{Performance Comparison with SOTA baselines}

Table~\ref{tab:SOTA_comparison} demonstrates that ConDiff achieves an accuracy of up to 83\% and an F1-score of 0.858, outperforming baselines by at least 3\% on both metrics while maintaining a balanced trade-off between sensitivity and PPV. 
Furthermore, ViT-based models achieved performance superior to CNN-based models. This may be due to their ability to use attention mechanisms to capture global dependencies between all input data elements. In contrast, CNN-based models, which apply uniform filters across the entire image, tend to focus on local content and may not be suited to capturing high inter-class similarity and intra-class variation.


\begin{figure}[!htbp]
  \centering
  \includegraphics[width=0.95\linewidth]{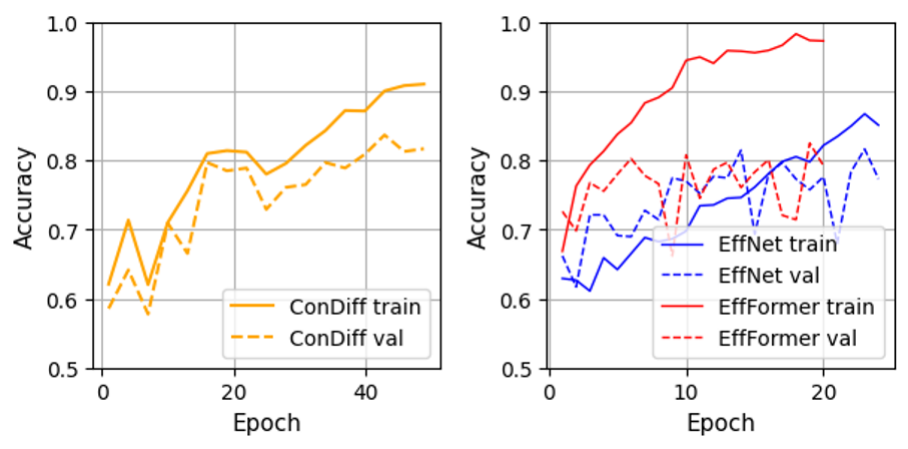}
  \caption{The learning accuracy trajectories of the ConDiff classifier and best-performing CNN-based (EfficientNet-B0) and ViT-based (EfficientFormer-L1) models, on the train and validation sets.} 
  \label{fig:learning_curve}
\end{figure}

However, since ViT-based models can be overfitted on our relatively small training dataset (illustrated in Fig.~\ref{fig:learning_curve}(Right)), they are not entirely suitable for infection classification. In contrast, the EfficientNet-B0 model is less susceptible to overfitting than the EfficientFormer-L1 model, which led us to select EfficientNet-B0 as the embedding network $f_\phi$ in the ConDiff classifier. Consequently, as shown in Fig.~\ref{fig:learning_curve}(Left), ConDiff effectively mitigates overfitting, which is attributed to the classifier's use of the Triplet loss function to learn to distinguish between similar and dissimilar images based on Euclidean distances in embedding space. 
The main ConDiff's trade-off is its computational time during inference. Table~\ref{tab:SOTA_comparison} shows that  ConDiff takes around 12-13 minutes to analyze the 251 test images using an NVIDIA A100 GPU while traditional classification models take less than 4 seconds on the same device. The longer inference time is because the ConDiff classifier iterates to synthesize the guided conditional images and measure their similarities to the input image.

\begin{figure}[!htbp]
  \centering
  
  \begin{subfigure}[b]{\columnwidth}
    \centering
    \includegraphics[width=0.9\linewidth]{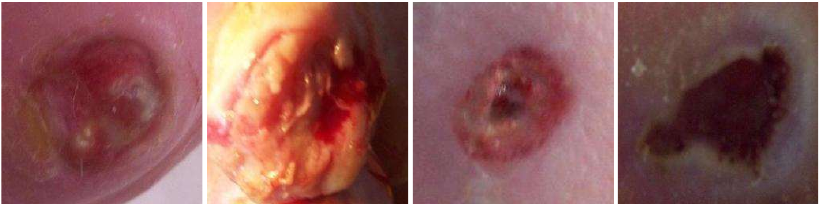}
    \caption{Uninfected DFUs  misclassified by ConDiff Classifier
}
  \end{subfigure}

  \begin{subfigure}[b]{\columnwidth}
    \centering
    \includegraphics[width=0.9\linewidth]{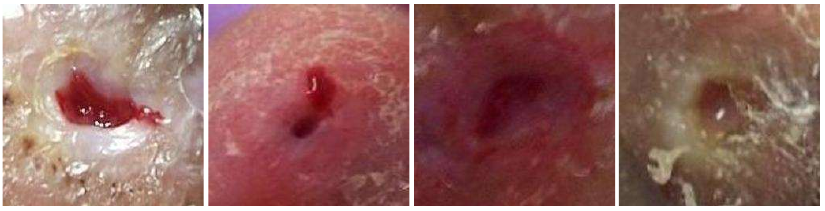}
    \caption{Infected DFUs  misclassified by ConDiff Classifier
}
  \end{subfigure}
  \caption{Examples of incorrectly classified DFU images for infection by our ConDiff Classifier.}
  \label{fig:error_analysis}
\end{figure}

\begin{figure*}[!htbp]
 \centering
 \minipage{0.49 \textwidth}
 \centering
     \includegraphics[width=0.99\textwidth]{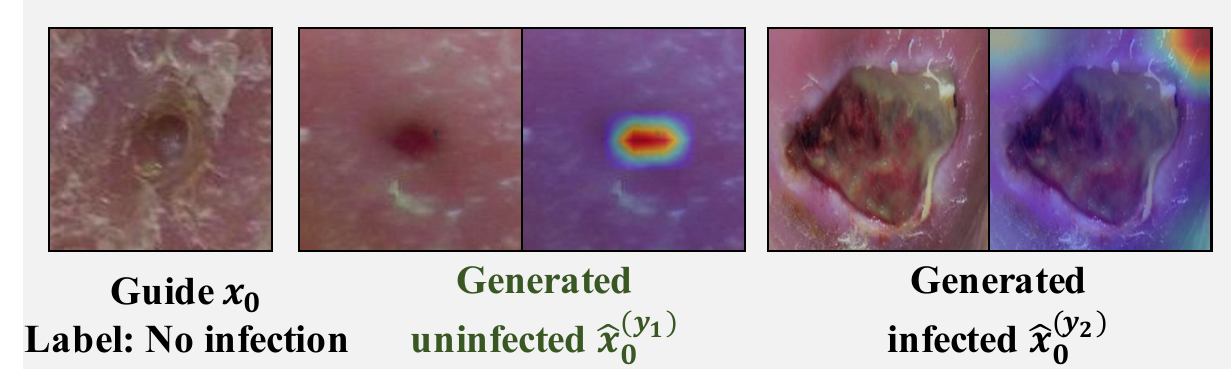}
     \subcaption{Correctly classified uninfected wound.}
 \endminipage 
 \minipage{0.49 \textwidth}
 \centering
    \includegraphics[width=0.99\textwidth]{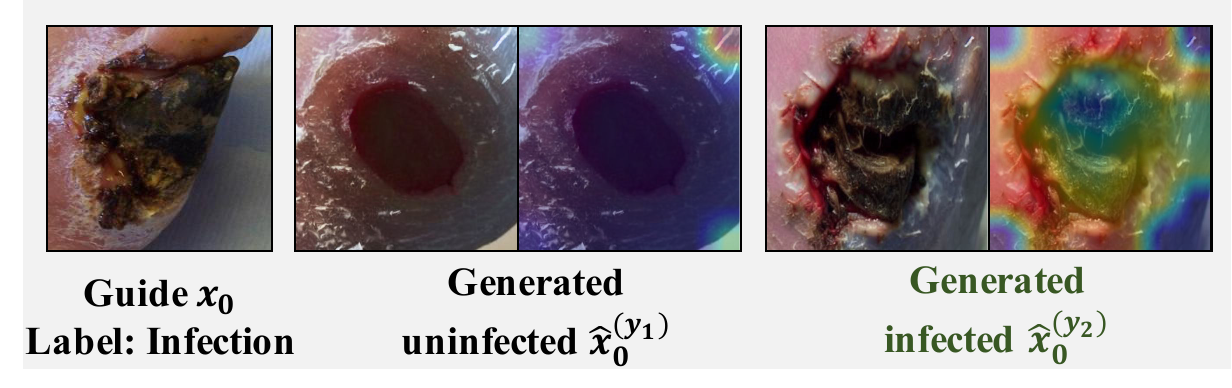}
    \subcaption{Correctly classified infected wound.}
 \endminipage \hfill 
  \minipage{0.48 \textwidth}
 \centering
    \includegraphics[width=0.99\textwidth]{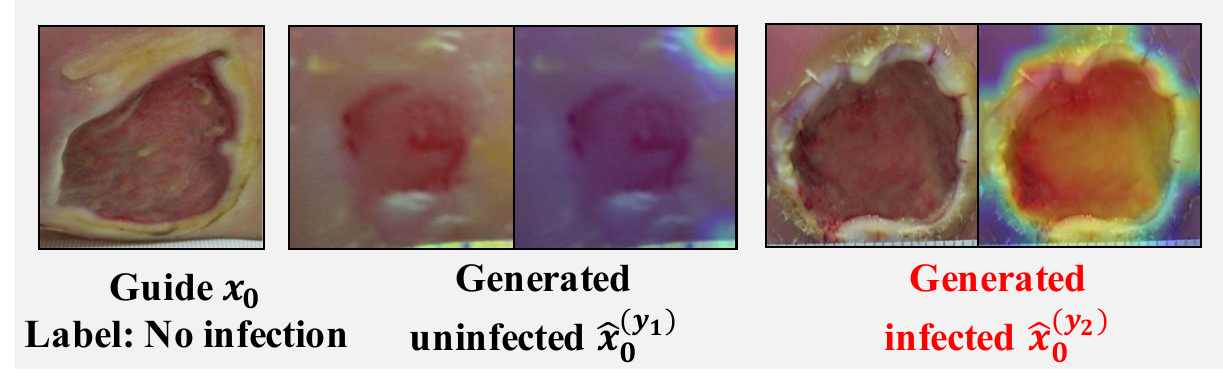}
    \subcaption{Misclassified uninfected wound as infected.}
 \endminipage
   \minipage{0.48 \textwidth}
 \centering
    \includegraphics[width=0.99\textwidth]{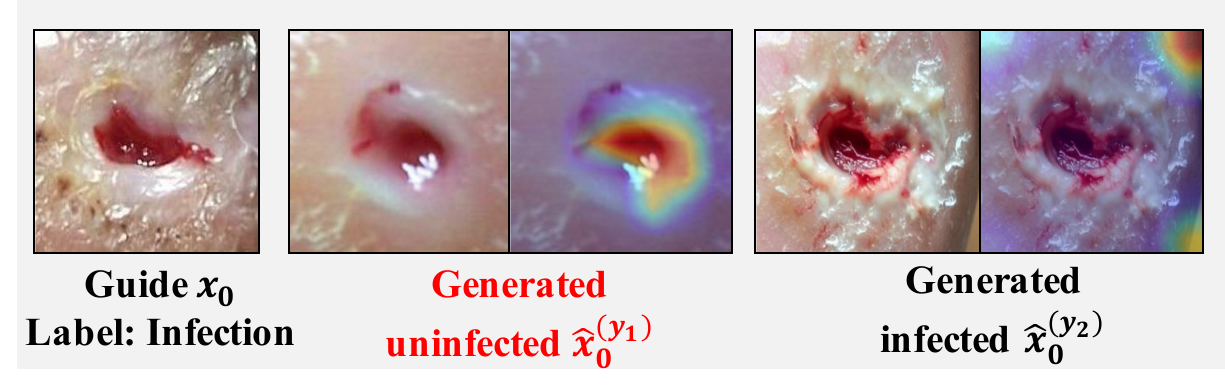}
    \subcaption{Misclassified infected wound as uninfected.}
 \endminipage

 \caption{Visualization of ConDiff predictions with corresponding Score-CAM images computed from the ConDiff's embedding model $f_\phi$. Each sub-figure shows an example with the guide image ($x_0$), conditional synthesized images representing uninfected $\hat{x}_0^{(y_1)}$ and infected $\hat{x}_0^{(y_2)}$ states, and their respective Score-CAM overlays indicate regions with similar features to $x_0$.}
 \label{fig:Score-CAM}
  \vspace{-2mm}
 \end{figure*}

\subsubsection{Exploring mis-classifications}

Fig.~\ref{fig:error_analysis}(a)-(b) show misclassified cases. The uninfected DFUs in Fig.~\ref{fig:error_analysis}(a) resemble infected wounds exhibiting characteristics such as a large reddish area or darkening of the wound, possibly caused by poor lighting conditions. The misclassified examples in Fig.~\ref{fig:error_analysis}(b) are due to the small size of the wounds and ambiguous features, such as a somewhat dried appearance, which confuses the infection classifier.

\subsubsection{Explaining image embedding similarity with Score-CAM} 




Fig.~\ref{fig:Score-CAM} presents Score-CAM visualizations that elucidate similarities the embedding model perceives between conditional synthesized images and their corresponding guide images $x_0$. Areas highlighted in red on the Score-CAM heatmaps shown in Fig.~\ref{fig:Score-CAM}, denote regions that the ConDiff classifier identified as having a high degree of similarity to $x_0$. For instance, Fig.~\ref{fig:Score-CAM}(a) illustrates that the classifier recognizes similar features in a synthesized uninfected image and the guide image, as indicated by the presence of a red spot in the heatmap. Conversely, the heatmap corresponding to the generated image conditioned on DFU infection does not reveal substantial similarity, except for a marginal overlap in the background at the top-right corner. Similarly, Fig.~\ref{fig:Score-CAM}(b) depicts accurate infection detection, where the embedding model $f_{\phi}$ concentrates on the necrotic tissue evident in both $x_0$ and $\hat{x}_0^{(y_2)}$. Nonetheless, the classifier is not infallible. Examples of misclassifications are demonstrated in Fig.~\ref{fig:Score-CAM}(c)-(d), where the embedding model incorrectly assesses a synthesized image conditioned on a different class as being more similar to the guide image, an error attributable to the high inter-class similarity in embedding space.
\begin{figure}[!htbp]
  \centering
  \includegraphics[width=0.8\linewidth]{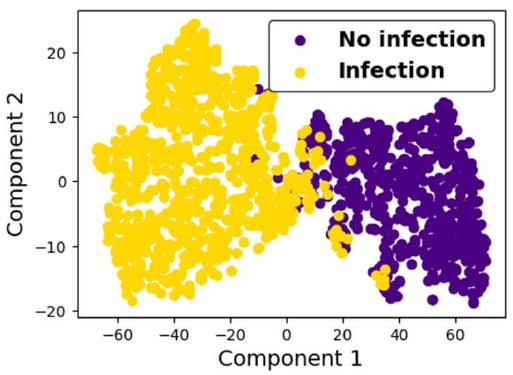}
  \caption{t-SNE plot of train image embedding computed by $f_\phi$.} 
  \label{fig:tsne}
  \vspace{-3mm}
\end{figure}

\begin{figure*}[!htbp]
\centering
  \includegraphics[width=0.9\textwidth]{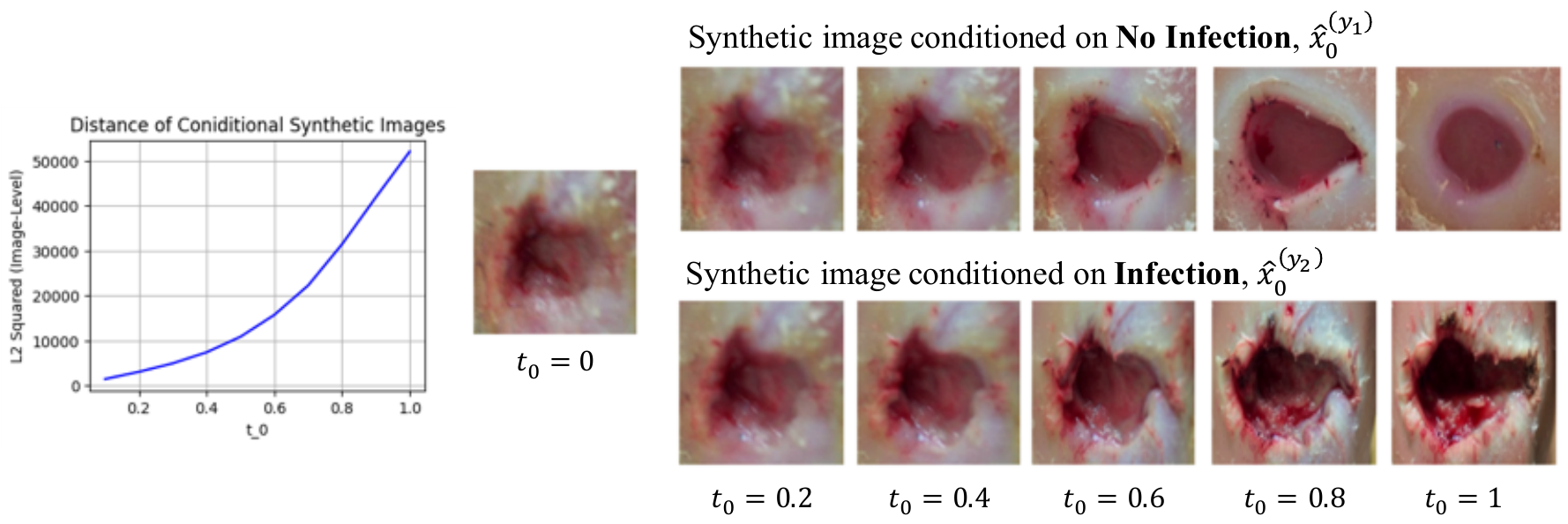}
  \caption{(Left) $L_2$ norm squared between $\hat{x}_0^{(y_1)}$ and $\hat{x}_0^{(y_2)}$ plot with respect to initial noise perturbation steps $t_0$, (Right) illustration of conditional synthesized images of ConDiff + CFG-DDIM sampling: $\omega=7.5$ with various $t_0$ initialization. As $t_0$ increases, the difference in image level between conditional generated images becomes larger.
}
  \label{fig:vary_t0}
  \vspace{-1.5mm}
\end{figure*}

\subsubsection{Visualizing image embedding} To further investigate the information gained from the embedding space, a t-SNE plot of train image embeddings is created, as shown in Fig.~\ref{fig:tsne}. It is observed that infected and uninfected wounds are separated into two distinct clusters. The Silhouette score of $0.586$ is relatively high, indicating well-separated clusters.

\subsection{Additional Experiments}


\subsubsection{Effects of different sampling methods on infection classification}
\label{sec:abla_samping}

We compared the DFU infection classification performance when two different sampling approaches were applied to ConDiff; (1) DDIM sampling and (2) CFG-DDIM sampling. Table.~\ref{tab:sampling_by_2_approches} shows that using CFG-DDIM sampling (Algorithm~\ref{alg_cfg_sampling}) significantly outperforms DDIM sampling in infection classification. This is because the guided generated images between 2 labels by DDIM are not different enough to make $D_\phi$ in Eq.~\ref{eq:d_phi} determine which of the synthesized images is most similar to the input image.
\begin{table}[!htbp]
\centering
\caption{Comparison of DFU infection classification by ConDiff Classifier with various sampling methods.}
\label{tab:sampling_by_2_approches}
\begin{tabular}{cccccc} 
\hline
Method & Acc & F1 & SEN & SPEC & PPV \\ \hline 
DDIM        & 0.603    & 0.635    & 0.581   & 0.641   & 0.699     \\
CFG-DDIM    & \textbf{0.833}    & \textbf{0.858}    & \textbf{0.858}   & \textbf{0.796}   & \textbf{0.858}     \\ \hline
\end{tabular}
\end{table}

\subsubsection{Effects of noise strengths $t_0$ on  infection classification}


Having demonstrated the efficacy of the CFG-DDIM reverse process in enabling the $D_\phi$ network to accurately predict infection status from a DFU image, this section focuses on experiments involving adding varying noise strengths $t_0$ to the input images. For these experiments, the guidance scale $\omega$, the control hyperparameter, was fixed at $7.5$.

\begin{table}[!htbp]
\centering
\caption{Quantitative comparison of the result of  DFU infection classification on the test data by ConDiff Classifier with different strengths $t_0$ of perturbed noise.}
\label{tab:vary_t0_perfomance}
\begin{tabular}{cccccc}
\hline
$t_0$ & Acc & F1 & SEN & SPEC & PPV \\  \hline
0.5      & 0.721     & 0.761      & 0.757     & 0.670     & 0.767        \\ 
0.6      & 0.721     & 0.757      & 0.736     & 0.699     & 0.779        \\ 
0.7      & 0.773     & 0.800      & 0.770     & 0.777     & 0.832        \\ 
0.8      & \textbf{0.833}   & \textbf{0.858}   & \textbf{0.858}     & 0.796   & 0.858 \\ 
0.9      & 0.809      & 0.829     & 0.791     & \textbf{0.835}      & \textbf{0.873}        \\ \hline
\end{tabular}
\end{table}

Table~\ref{tab:vary_t0_perfomance} highlights the significant role of perturbed noise in image synthesis. As the noise strength $t_0$ increases, the guided synthesized images corresponding to different labels exhibit greater divergence, facilitating more straightforward predictions by our distance-based classifier $D_\phi$. However, when $t_0$ is increased beyond certain thresholds, the distances between input images and their respective guided synthesized counterparts become excessively large for both labels. This increase in distance diminishes the distinguishability of $D_\phi$. Classification with $t_0$ values ranging from $0.1$ to $0.4$ was excluded from our analysis, as depicted in Fig.~\ref{fig:vary_t0} (Left), where it is shown that the average difference in the $L_2$ norm squared between the two guided synthetic images $\hat{x}_0^{(y_1)}$ and $\hat{x}_0^{(y_2)}$ is relatively small. Fig.~\ref{fig:vary_t0} (Right) presents examples of conditional synthetic images generated by ConDiff + CFG-DDIM sampling across different $t_0$ values.

\begin{figure*}[!ht]
  \centering
  \begin{subfigure}[b]{\textwidth}
  \centering
    \includegraphics[width=0.9\linewidth]{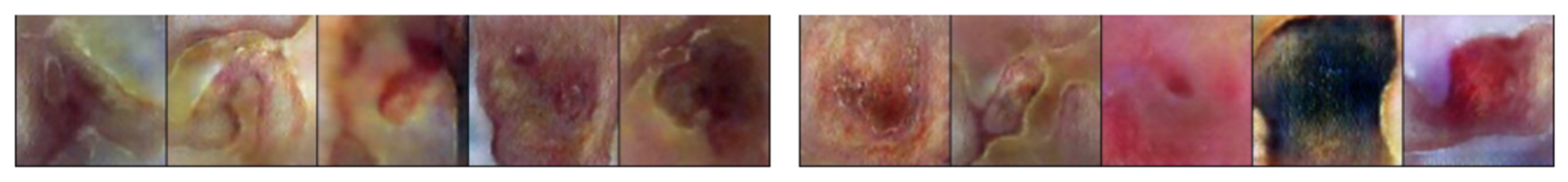}
    \caption{cGAN generation: $FID=11.201$, $IS=2.966$}
    \label{fig:a}
  \end{subfigure}

  \begin{subfigure}[b]{\textwidth}
  \centering
    \includegraphics[width=0.9\linewidth]{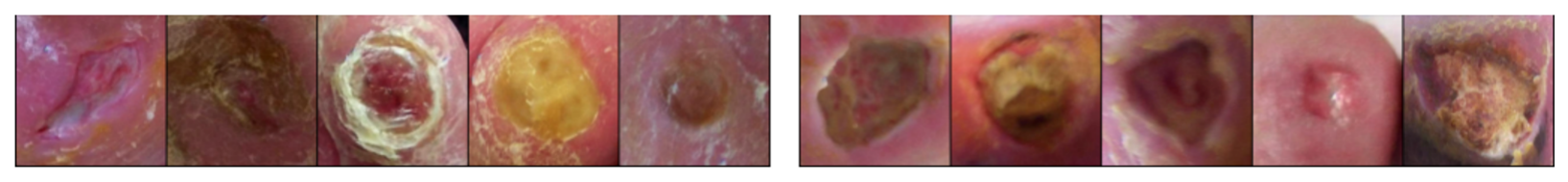}
    \caption{ConDiff, DDIM sampling method: $FID=2.917$, $IS=3.662$}
    \label{fig:b}
  \end{subfigure}

  \begin{subfigure}[b]{\textwidth}
  \centering
    \includegraphics[width=0.9\linewidth]{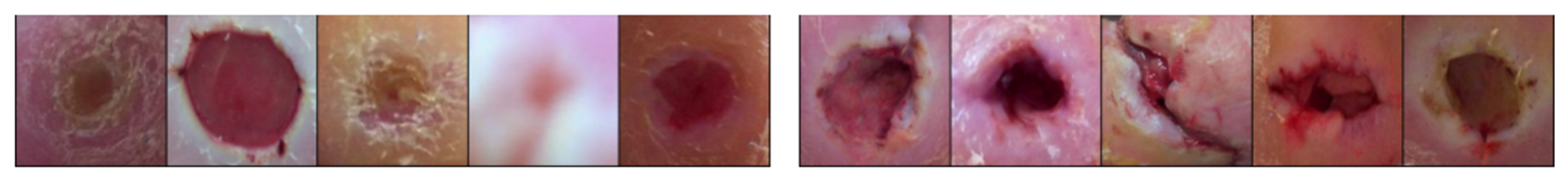}
    \caption{ConDiff, CFG-DDIM sampling method with $\omega=3$: $FID=3.779$, $IS=3.669$ }
    \label{fig:c}
  \end{subfigure}

  \begin{subfigure}[b]{\textwidth}
  \centering
    \includegraphics[width=0.9\linewidth]{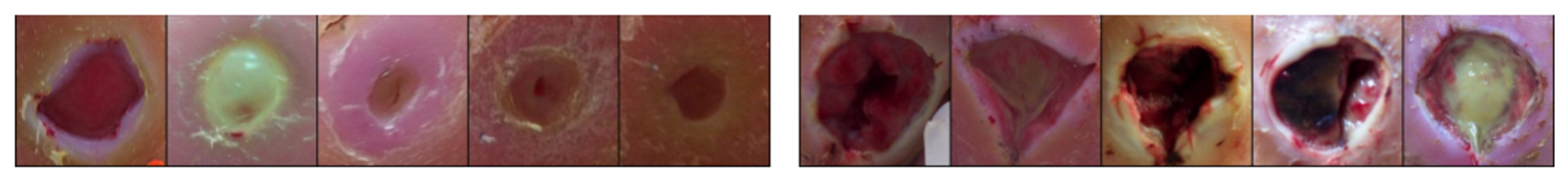}
    \caption{ConDiff, CFG-DDIM sampling method with $\omega=7.5$: $FID=5.068$, $IS=3.885$ }
    \label{fig:d}
  \end{subfigure}
  \caption{Guided Image Generation. Condition on the label: (Left) uninfected wound and (Right) infected wound.}
  \label{fig:generate_images}
\end{figure*}

\begin{table}[!htbp]
\centering
\caption{Quality measurement of conditional synthesized images by generative models.}
\label{tab:FID_gen}
\begin{tabular}{lccc}
\hline
\multicolumn{1}{c}{Model}    & FID Score $\downarrow$      & IS $\uparrow$  \\ \hline
ConDiff + DDIM        & \textbf{2.917}              & 3.662          \\ 
ConDiff + CFG-DDIM, $\omega=3$   &  3.779             & 3.669      \\
ConDiff + CFG-DDIM, $\omega=7.5$   &  5.068             & \textbf{3.885}  \\ 
Conditional GAN \cite{cGAN}       & 11.201       & 2.965           \\ 
\hline
\end{tabular}
\end{table}

\subsubsection{Evaluation of Conditional Image Synthesis}


Previously, in Sec.~\ref{sec:abla_samping}, it was established that CFG-DDIM sampling is more effective for distance-based classification. Nevertheless, realistic DFU images can still be generated using the DDIM sampling method. 
Table~\ref{tab:FID_gen} reveals that image synthesis using ConDiff + DDIM results in the lowest FID score, indicating that the distribution of synthesized images closely resembles that of real data. However, the Inception Score (IS) for the ConDiff + DDIM approach is lower than the ConDiff + CFG-DDIM approach. This finding aligns with Ho et al.'s experiment \cite{CFG}, which highlights a trade-off between FID and IS. In our context, the IS reflects the ease of differentiating between conditional synthesized images. Consequently, the ConDiff Classifier using the DDIM sampling approach underperforms relative to the CFG-DDIM sampling approach, as seen in Table~\ref{tab:sampling_by_2_approches}. This is attributed to the challenge in label clarification due to the lower IS. Note that IS is not significantly different across our diffusion models because the IS metric considers not only the clarity but also the diversity of synthesized images, as shown in Fig.~\ref{fig:generate_images}. Additionally,  the experiment employed the popular conditional GAN approach \cite{cGAN} for generating DFU images conditioned on infection status, providing a comparison to diffusion methods. 
The results indicate that the quality of synthesized images using the conditional GAN is inferior to those produced by diffusion methods.

\section{Discussion}
\label{sec:Discussion}
\underline{Summary of Findings}:
\textbf{The ConDiff classifier outperforms other deep learning models in detecting infections in DFU images} by minimizing Triplet loss instead of binary cross-entropy, enabling it to effectively match input images with the most similar conditionally synthesized images in embedding space.

\textbf{Overfitting Mitigation:} ConDiff's training strategy in \textit{Training Stage 2}, involving triplet loss, not only enhances its performance but also reduces overfitting by learning to discern between infected and uninfected wounds in the dataset. 

\textbf{CFG-DDIM Integration:} When applying CFG-DDIM for the sampling process, ConDiff outperforms the standard DDIM method in infection classification. This is significant because, despite DDIM's proven efficiency in generating DFU-like images, CFG-DDIM's guidance-label-influenced image generation aligns better with ConDiff's distance-based classification.

\textbf{Score-CAM enhances ConDiff's Interpretability:} As shown in Figure~\ref{fig:Score-CAM}, ConDiff focuses on wound features critical for accurately predicting infection status in its decision-making.

\underline{Limitation}: ConDiff's major drawback is its high computational cost during inference, taking 3-4 seconds per image on an NVIDIA A100 GPU, in contrast to less than 0.02 seconds for other models. This is due to its generative discriminative approach, which synthesizes conditional images for each input by gradually removing noise through a reverse diffusion process. The Stability AI research group has recently found a way to overcome this limitation by reducing the number of sampling steps to just one step when generating conditional images by leveraging the adversarial diffusion distillation technique \cite{sdxl-turbo2023}.

\underline{Future Work}: The potential of the embedding network $f_\phi$ in transforming DFU images into meaningful vectors suggests future research in more efficient prediction approaches. Exploring multi-modal data, including thermal images or generative medical notes from Large Language Models (LLMs), could enhance the classification capabilities of deep-learning models. The ConDiff classifier might also be applied to other medical imaging tasks and other wound types, such as classifying pressure injury severity \cite{ay2022deep}.

\section{Conclusion}
\label{sec:conclusion}

This study introduced the Guided Conditional Diffusion classifier (ConDiff), a new framework for classifying Diabetic Foot Ulcer (DFU) infections from wound images. Outperforming traditional models by at least 3\%, ConDiff achieves up to 83\% accuracy and an F1-score of 0.858. Its unique approach utilizes Triplet loss instead of standard cross-entropy minimization, enhancing robustness and reducing overfitting. This is especially important in medical imaging where datasets are often small. ConDiff employs a forward diffusion process, to add a specific amount of Gaussian noise into input images, and a reverse diffusion with classifier-free guidance to iteratively refine these images for classification based on the closest Euclidean distance in an embedding space. The ConDiff's effectiveness suggests significant potential in improving DFU management, particularly in regions with limited medical resources. Its precise, real-time infection detection could play a crucial role in early DFU infection identification, reducing serious complications such as limb amputation.

\section*{Acknowledgment}
This research was performed using computational resources supported by the Academic \& Research Computing group at Worcester Polytechnic Institute.

\bibliographystyle{IEEEtran}
\bibliography{main} 

\end{document}